\documentclass{article}

\usepackage[final]{neurips_2019}

\usepackage[utf8]{inputenc} 
\usepackage[T1]{fontenc}    
\usepackage{hyperref}       
\usepackage{url}            
\usepackage{booktabs}       
\usepackage{amsfonts}       
\usepackage{nicefrac}       
\usepackage{microtype}      

\usepackage{graphicx}
\usepackage{enumitem}
\usepackage{xcolor}
\usepackage{amsmath}

\newcommand\link[1]{\textcolor{blue}{#1}}

\newcommand{\norm}[1]{\left\lVert#1\right\rVert}

\title{LucidDream: Controlled Temporally-Consistent DeepDream on Videos}
  
\author{
Joel Ruben Antony Moniz, Eunsu Kang, Barnabás Póczos \\
Carnegie Mellon University
}
 
\begin{document}

\maketitle

\begin{abstract}
In this work, we aim to propose a set of techniques to improve the controllability and aesthetic appeal when DeepDream, which uses a pre-trained neural network to modify images by hallucinating objects into them, is applied to videos. In particular, we demonstrate a simple modification that improves control over the class of object that DeepDream is induced to hallucinate. We also show that the flickering artifacts which frequently appear when DeepDream is applied on videos can be mitigated by the use of an additional temporal consistency loss term.
\end{abstract}

\section{Introduction}

Digital artists have increasingly adopted techniques that use deep learning as part of their repertoire. One such technique is DeepDream \citep{DeepDream}, which uses a pre-trained convolutional architecture and updates the image to force the network to "hallucinate" objects. While this technique has been perfected for images, a popular way of applying DeepDream to videos is to directly apply it frame-by-frame. A significant drawback of this approach is that it causes a flickering effect due to the lack of temporal consistency between frames, often detracting from the overall aesthetic appeal of the end result. Another drawback of DeepDream is that controlling the objects hallucinated in the input image is often done by trial-and-error, and is thus not straightforward.

In this work, we describe two simple modifications to the traditional DeepDream formulation to improve its controllability and applicability to videos. The first enhances the degree of control when applying DeepDream to images by improving the ability to hallucinate specific classes by updating the image to maximize the network's final classification layer's logits, as opposed to the intermediate convolutional layers. The second improves DeepDream's applicability to videos, resolving the  flickering issue by drawing inspiration from recent work in style transfer  \citep{styleTransfer,styleTransferVideo} and leveraging temporal consistency loss terms. 

\begin{figure*}[!htb]
  \centering
    \includegraphics[width=.32\textwidth,height=\textheight,keepaspectratio]{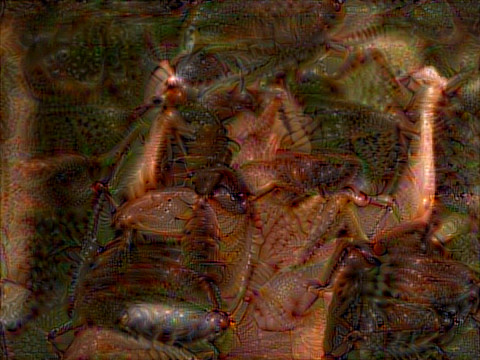}
    \includegraphics[width=.32\textwidth,height=\textheight,keepaspectratio]{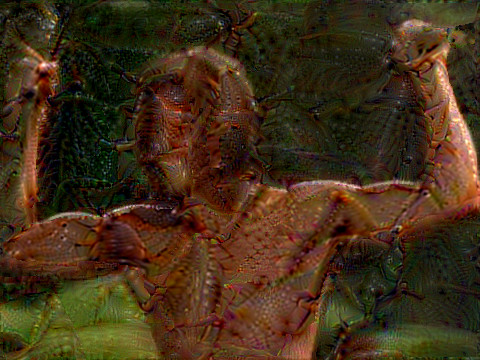}
    \includegraphics[width=.32\textwidth,height=\textheight,keepaspectratio]{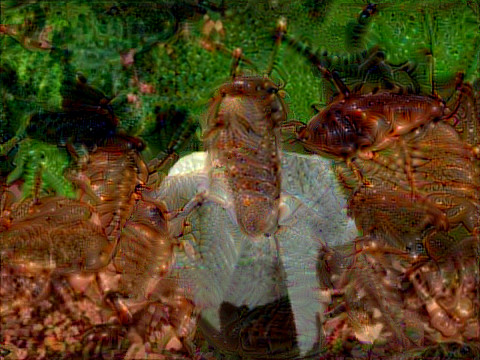}
    \includegraphics[width=.32\textwidth,height=\textheight,keepaspectratio]{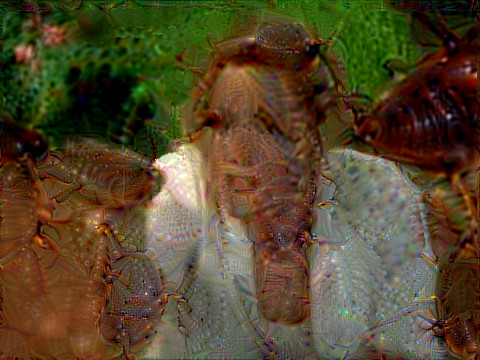}
    \includegraphics[width=.32\textwidth,height=\textheight,keepaspectratio]{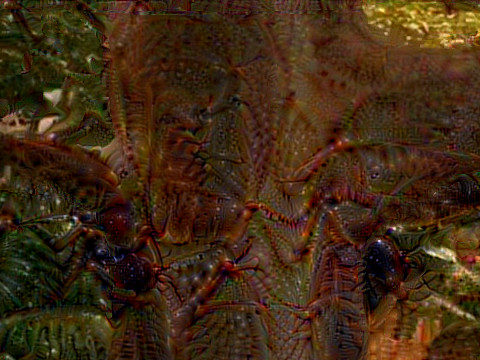}
    \includegraphics[width=.32\textwidth,height=\textheight,keepaspectratio]{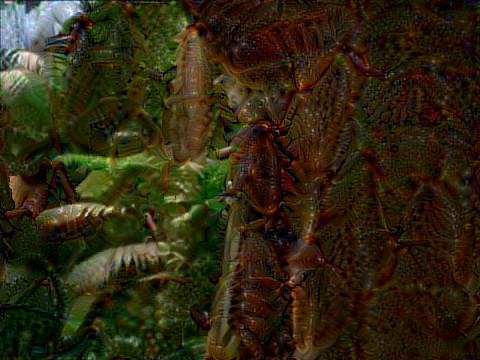}
  \caption{Controlled DeepDream with temporal consistency and scene change detection applied on frames (from top-left) 1, 25, 50, 75, 100 and 120 in a 120 frame clip of size $480\times360$ with 3 continuous takes. The \texttt{cockroach} ImageNet class hallucinated. Best viewed zoomed in and in colour. The complete video is available \link{\href{https://drive.google.com/open?id=1SkYCJK4tmkZW7GzrMTQNt2PTkhCSz8T7}{here}}.}
  \label{fig:george}
\end{figure*}

\section{Method}

\textbf{DeepDream} The original DeepDream framework works with a model of the Inception \citep{Inception} architecture fully trained on ImageNet \citep{imagenet}. The loss is defined as $\mathcal{L}_{l, a} = -\norm{\mathcal{F}_{l,a}(\mathcal{I})}^2_F$, where the function $\mathcal{F}_{l, a}(\mathcal{I})$ yields the output of the $l^{th}$ layer's $a^{th}$ feature map after applying the network to the input image $\mathcal{I}$, and $\norm{.}^2_F$ represents the squared Frobenius norm. The original image is then updated to minimize this loss, for example, using gradient descent.

\textbf{Controlled DeepDream} To control DeepDream, we propose maximizing square of the pre-softmax activations (i.e., the logits) $\mathcal{F}_{c}(\mathcal{I})$ of the class $c$ to be hallucinated, as opposed to using an intermediate feature map. Correspondingly, the loss to be minimized becomes $\mathcal{L}_{c} = - \mathcal{F}_{c}(\mathcal{I})^2$.

A drawback of this formulation is that the loss is no longer applied to a fully convolutional network- the input image must be of size 224x224. To overcome this, we apply controlled DeepDream on uniformly randomly sampled tiles of size 224x224: we describe this tiling in detail in Section \ref{sec:tile}.

\textbf{Temporal Consistency in DeepDream} 

Applying DeepDream individually to each frame in a video causes a flickering effect. This occurs because of the lack of a constraint forcing adjacent frames to be consistent with each other when patterns and objects are hallucinated. Inspired by recent work on applying neural style transfer to videos, we propose the application of similar short-term and long-term temporal consistency losses. As part of this process, we find that certain hyperparameter sets yield artistically interesting and visually distinct results: we describe these in Section \ref{sec:altvisual}. The losses are similar to those used when applying style transfer to videos \citep{styleTransferVideo}, and are described below.

Given the $i^{th}$ frame and the $i-j^{th}$ frame, let $\mathbf{x^{(i)}}$ be the output image, let $\mathbf{w^{(i-j, i)}}$ be $\mathbf{x^{(i-j)}}$ mapped to  $\mathbf{x^{(i)}}$ using the optical flow between the two input frames. Further, let $\mathbf{c^{(i-j, i)}}$ represent the temporal consistency between the two frames (in the form of a boolean mask), indicating the presence of de-occlusions and motion boundaries, as described in \citep{styleTransferVideo,sundaram2010dense}.

The \textbf{long-term temporal consistency loss} is:

\begin{align*}
    \mathcal{L}_{lt} &= \frac{1}{D}  \sum_{j\in J : i-j\geq1} \sum_{k=1}^{D} c_{l}^{(i-j, i)}[k] . (x^{(i)}[k] - w^{(i-j, i)}[k])^2 \\
     \textbf{c}_{l}^{(i-j, i)} &= max \left(  \textbf{c}^{(i-j, i)} - \sum_{k\in J : i-k > i-j} \textbf{c}^{(i-k, i)}, \textbf{0}  \right)
\end{align*}

where $D$ is the dimensionality of the input image and J is the set indices to be subtracted from the current frame index $i$, i.e., the set of previous frames to be taken into account. The \textbf{short-term temporal consistency loss} $\mathcal{L}_{st}$ is obtained by setting $J=\{1\}$.

\textbf{LucidDream} The final LucidDream update, which yields temporally consistent videos and improves controllability of the hallucinated class is then given by $\mathcal{L} = \alpha \mathcal{L}_c + \beta \mathcal{L}_{st} + \gamma \mathcal{L}_{lt}$, where $\alpha$, $\beta$ and $\gamma$ are hyperparameters. Each frame in the video is then updated to minimize this loss term.

\bibliography{main_neurips2019}
\bibliographystyle{icml2018}

\clearpage

\appendix

\section{Appendix: Experimental Details and Additional Visualizations}

\begin{figure*}[!htb]
  \centering
    \includegraphics[width=.32\textwidth,height=\textheight,keepaspectratio]{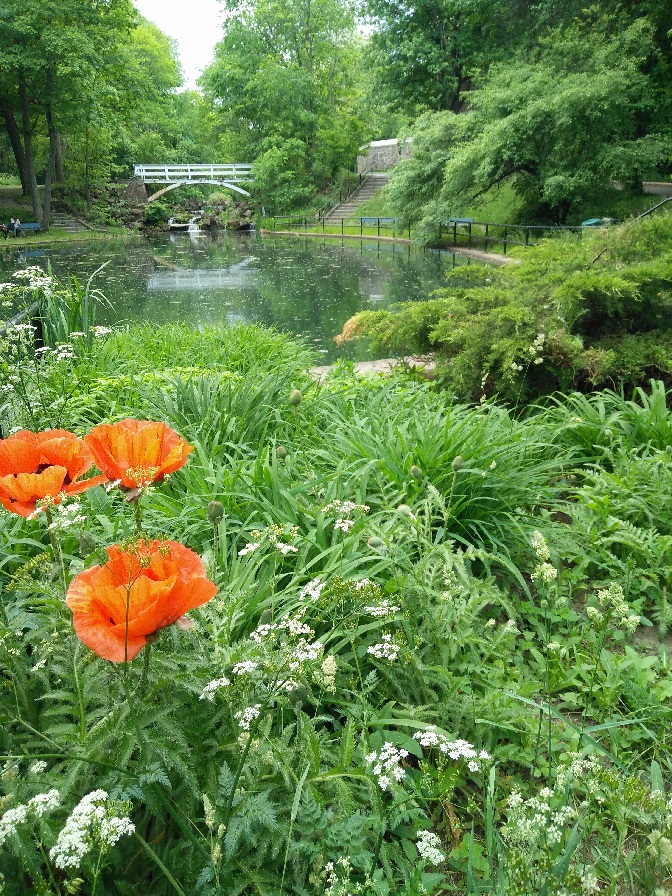}
    \includegraphics[width=.32\textwidth,height=\textheight,keepaspectratio]{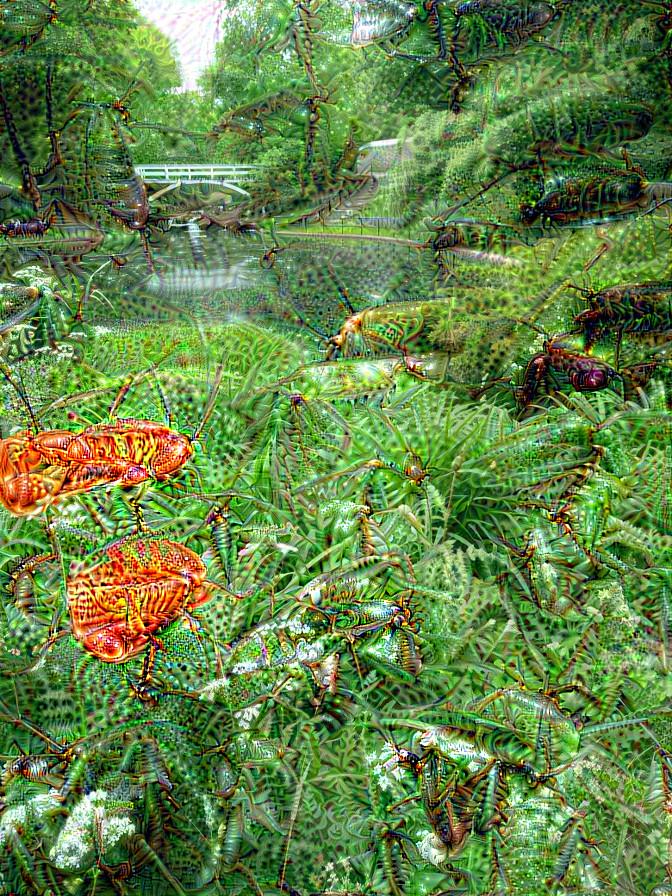}
    \includegraphics[width=.32\textwidth,height=\textheight,keepaspectratio]{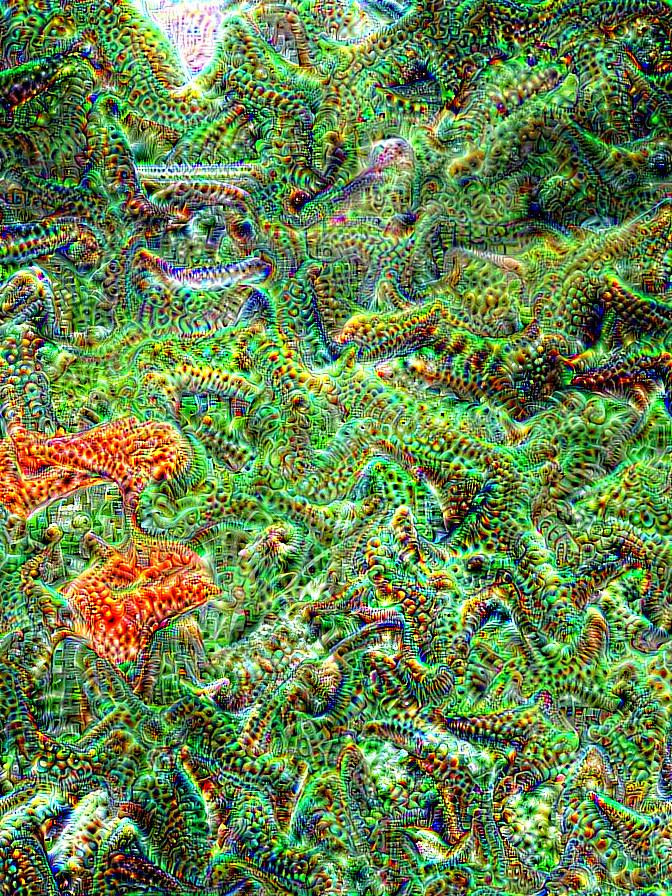}
  \caption{Left to right: original image, image with \texttt{cockroach} ImageNet class hallucinated, and image with \texttt{starfish} ImageNet class hallucinated. All images of original size $672\times896$ (i.e., both dimensions multiples of $224$).}
  \label{fig:flower}
\end{figure*}

\begin{figure*}[!htb]
  \centering
    \includegraphics[width=.49\textwidth,height=\textheight,keepaspectratio]{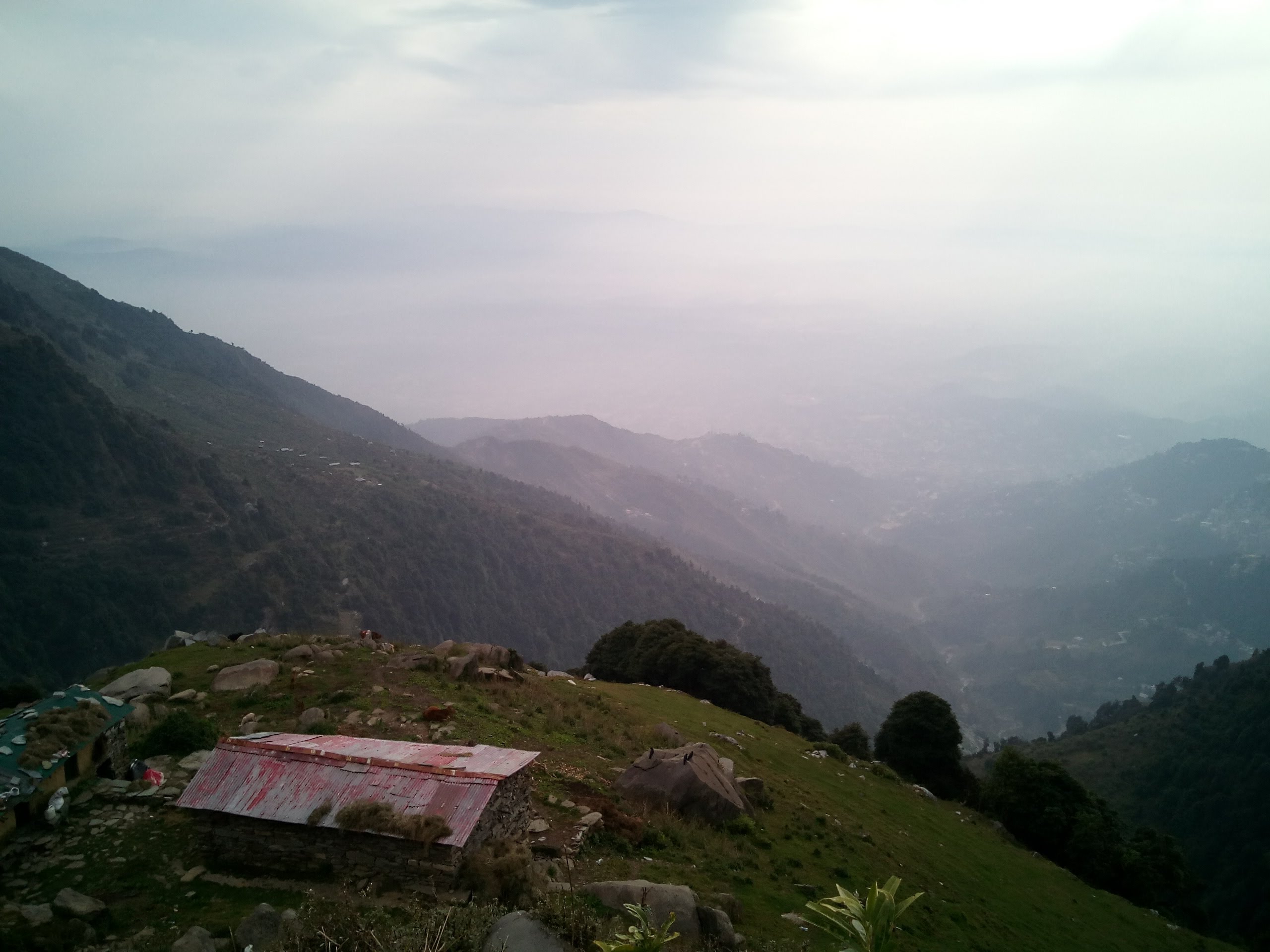}
    \includegraphics[width=.49\textwidth,height=\textheight,keepaspectratio]{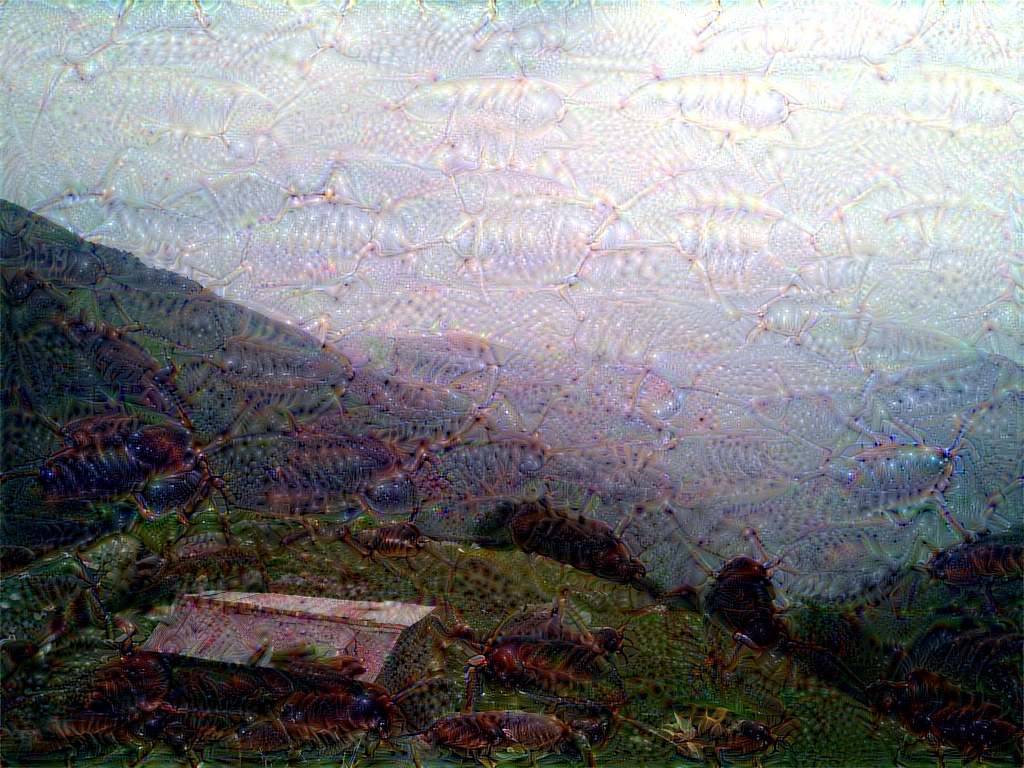}
    \includegraphics[width=.49\textwidth,height=\textheight,keepaspectratio]{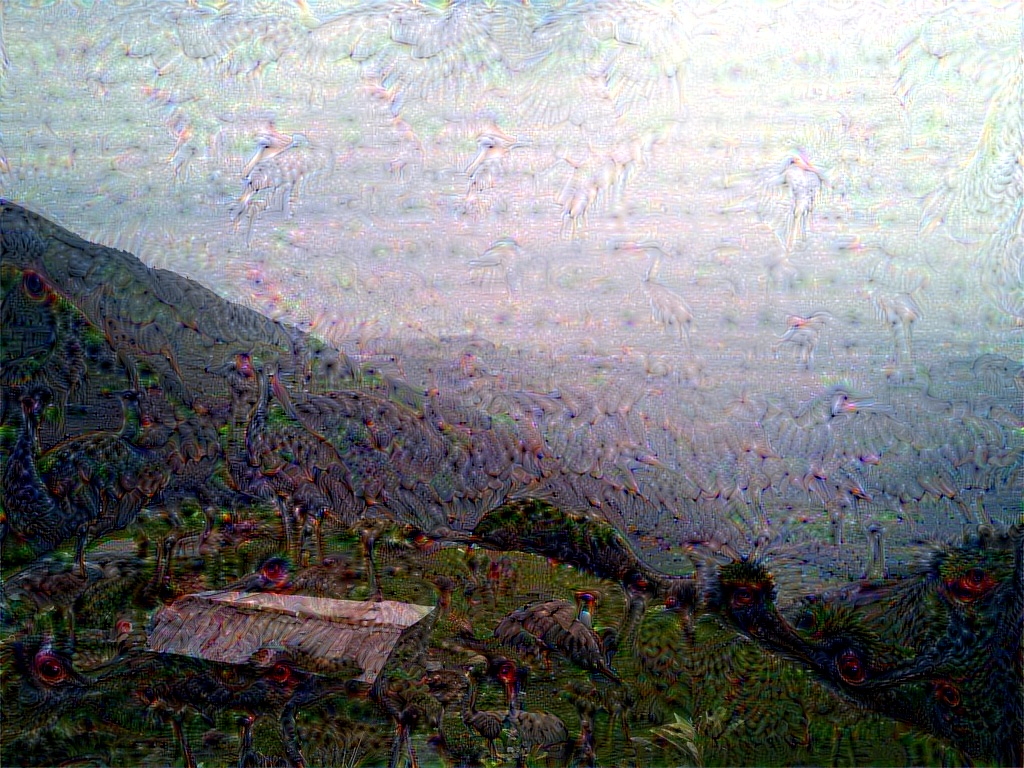}
    \includegraphics[width=.49\textwidth,height=\textheight,keepaspectratio]{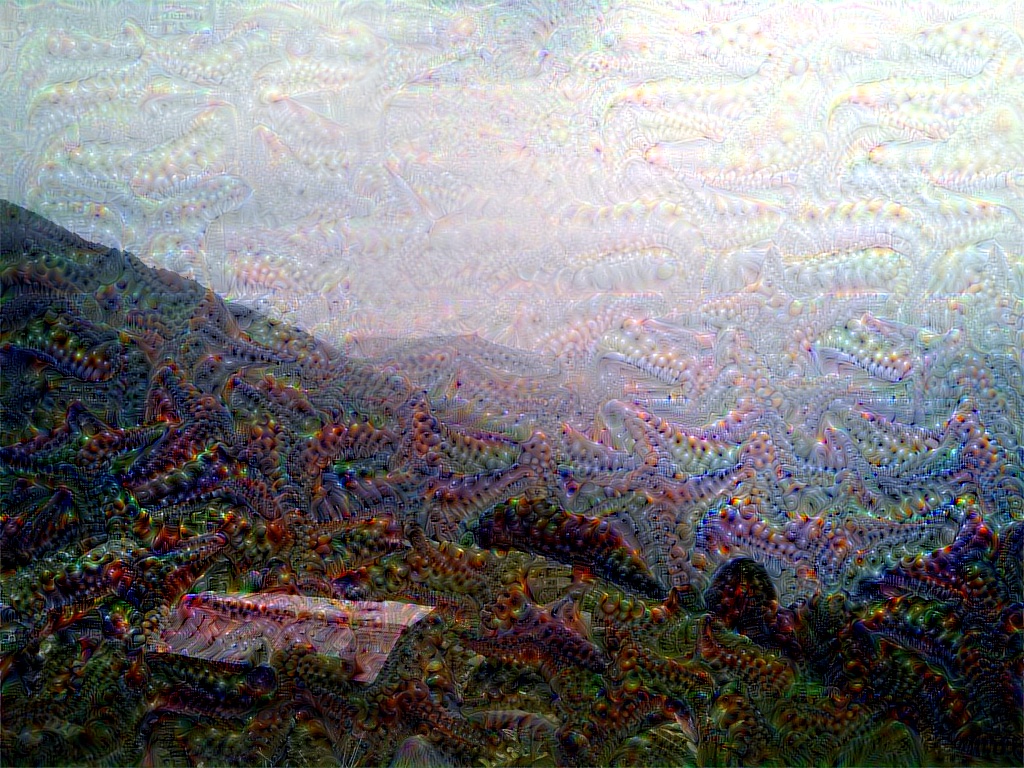}
  \caption{Clockwise from the top-left: original image, \texttt{cockroach} ImageNet class hallucinated, \texttt{starfish} ImageNet class hallucinated, and \texttt{crane} ImageNet class hallucinated. Best viewed zoomed in and in colour. All images of original size $2560\times1920$ (i.e., neither dimension a multiple of $224$).}
  \label{fig:mcleod}
\end{figure*}

\subsection{Tiling} \label{sec:tile}

As discussed, one drawback of using logits to control the hallucinated class directly is that the network used in the loss optimmization is no longer fully convolutional. To overcome this drawback, we use a randomized tiling mechanism. In particular, a random starting point is chosen in the image. The image is then rolled circularly along both the vertical and horizontal axes so as to make the random starting point the origin and divided up into 224x224 tiles (224 being the input size that most convnet architectures pre-trained on ImageNet \citep{imagenet} commonly take as input). Controlled DeepDream is then applied to each such 224x224 tile in the image. Note that while doing this, the boundary region along the width and height axes will be left as a "margin", with DeepDream not applied to that region. However, this region will be small, with the smaller dimension of the rectangular margin along the width ($w$) and height ($h$) dimensions respectively being $w\%224$ and $h\%224$. Further, this margin effectively gets "smoothed" out: because the starting point is randomly chosen, all regions in the image end up having a uniform probability of being excluded for a specific iteration, and the number of times the controlled DeepDream is applied to each region is the same in expectation; the random selection of the origin thus avoids tiling artifacts from arising. While \citep{deepdream_tf} used a similar tiling and jitter mechanism, the primary goal of the tiling was for memory efficiency, with only a small amount of jitter applied to visually improve the final output and avoid tile seams; here, however, the jitter size can be thought of as the entire image (as opposed to a small amount), and the tiling is crucial to overcome the network no longer being fully convolutional.

\subsection{Over-hallucination} \label{sec:overhall}

\begin{figure*}[!htb]
  \centering
    \includegraphics[width=.32\textwidth,height=\textheight,keepaspectratio]{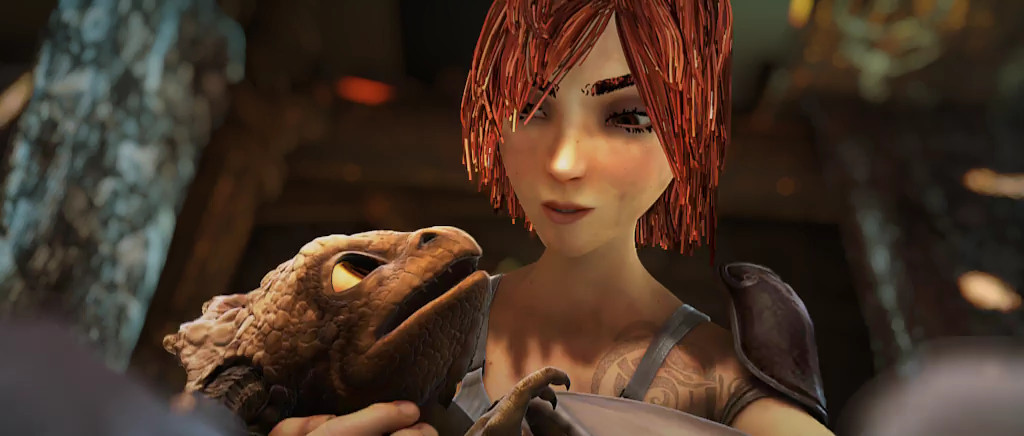}
    \includegraphics[width=.32\textwidth,height=\textheight,keepaspectratio]{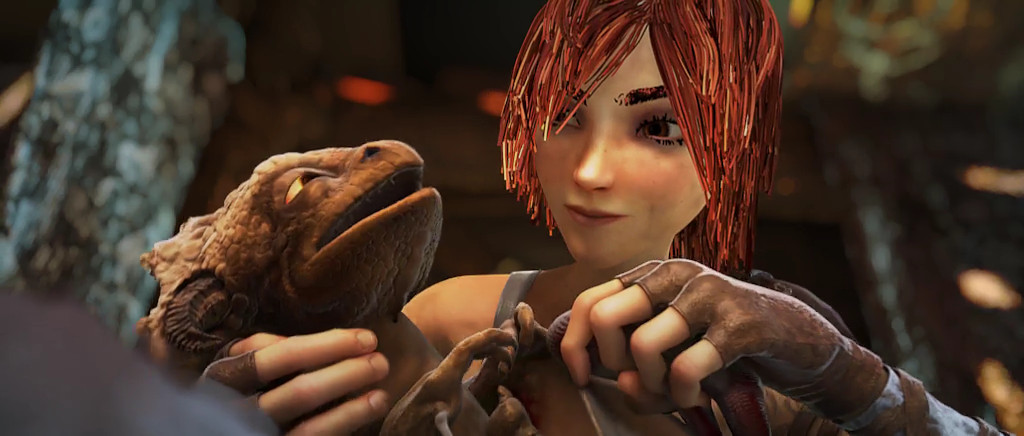}
    \includegraphics[width=.32\textwidth,height=\textheight,keepaspectratio]{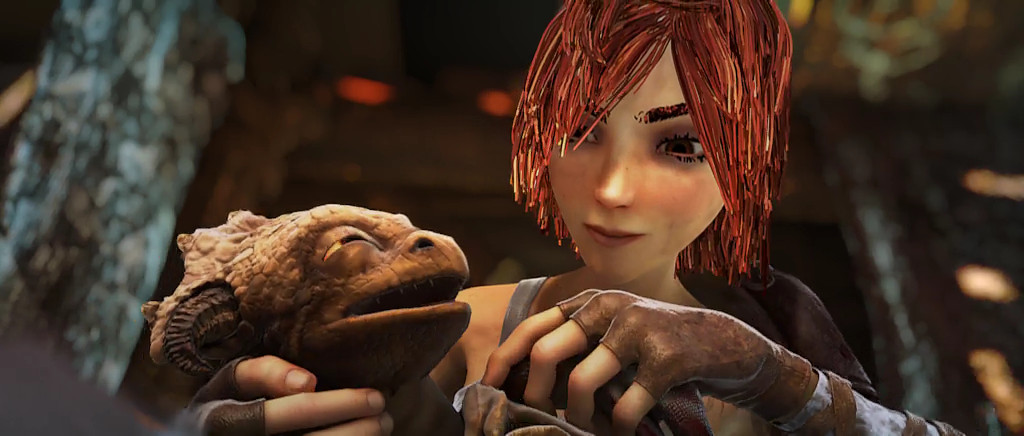}
    \includegraphics[width=.32\textwidth,height=\textheight,keepaspectratio]{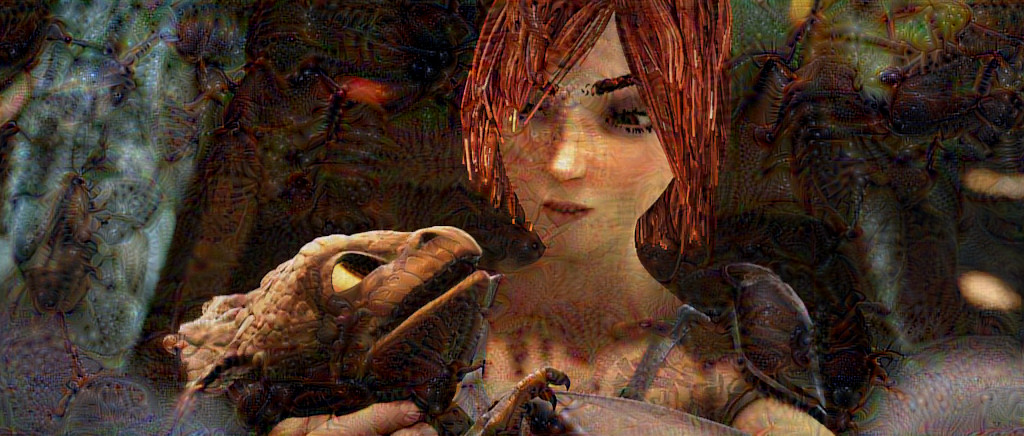}
    \includegraphics[width=.32\textwidth,height=\textheight,keepaspectratio]{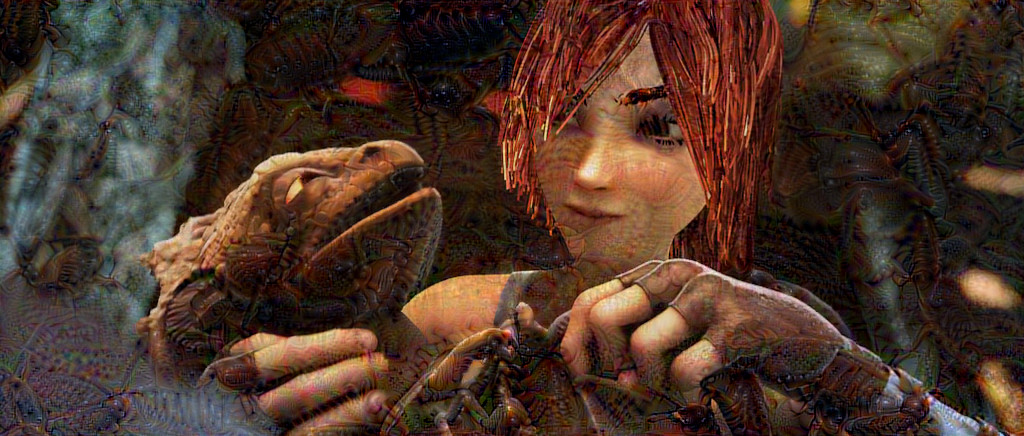}
    \includegraphics[width=.32\textwidth,height=\textheight,keepaspectratio]{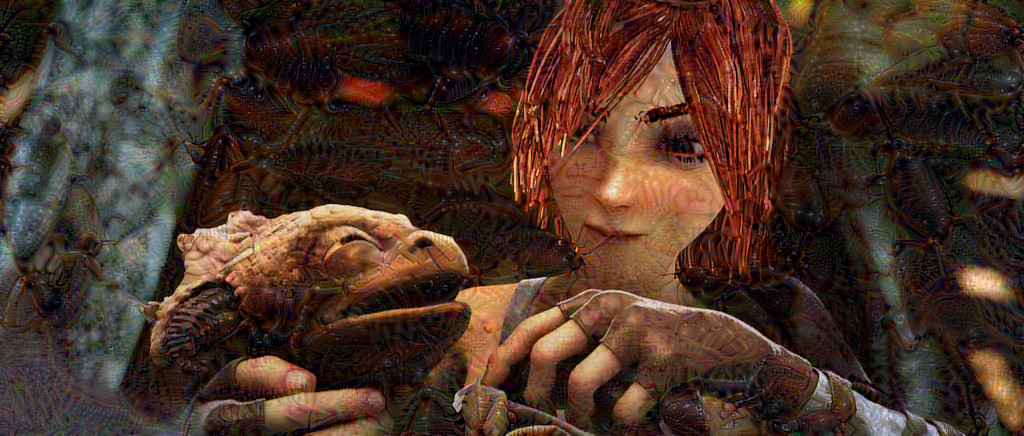}
    \includegraphics[width=.32\textwidth,height=\textheight,keepaspectratio]{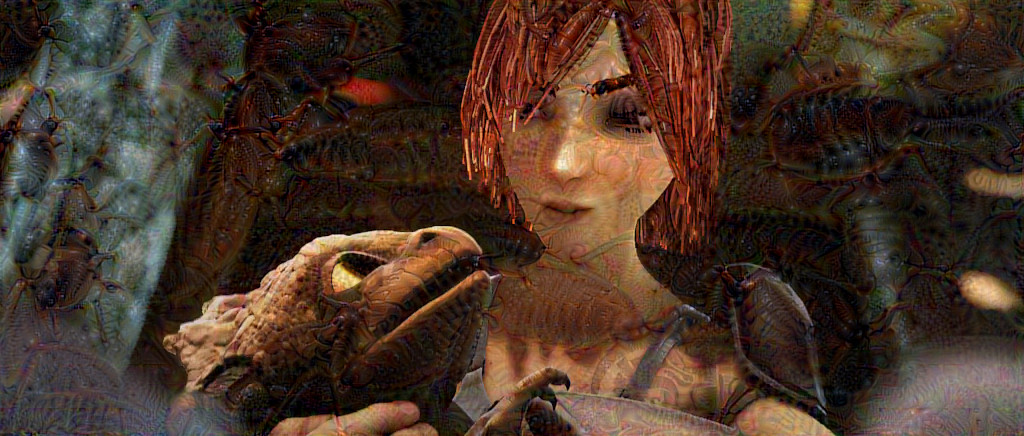}
    \includegraphics[width=.32\textwidth,height=\textheight,keepaspectratio]{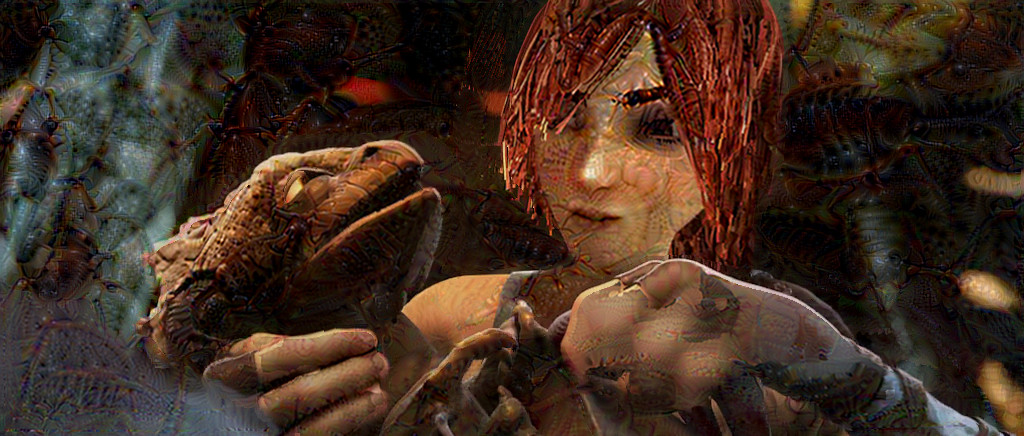}
    \includegraphics[width=.32\textwidth,height=\textheight,keepaspectratio]{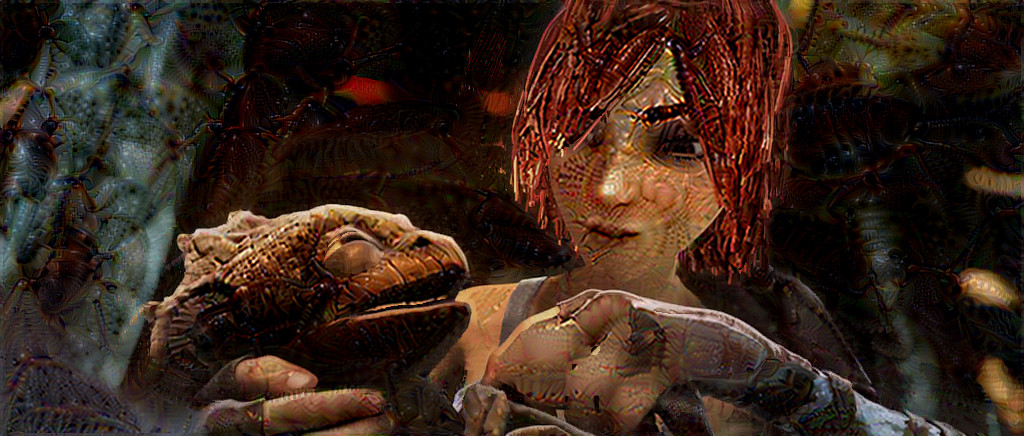}
    \includegraphics[width=.32\textwidth,height=\textheight,keepaspectratio]{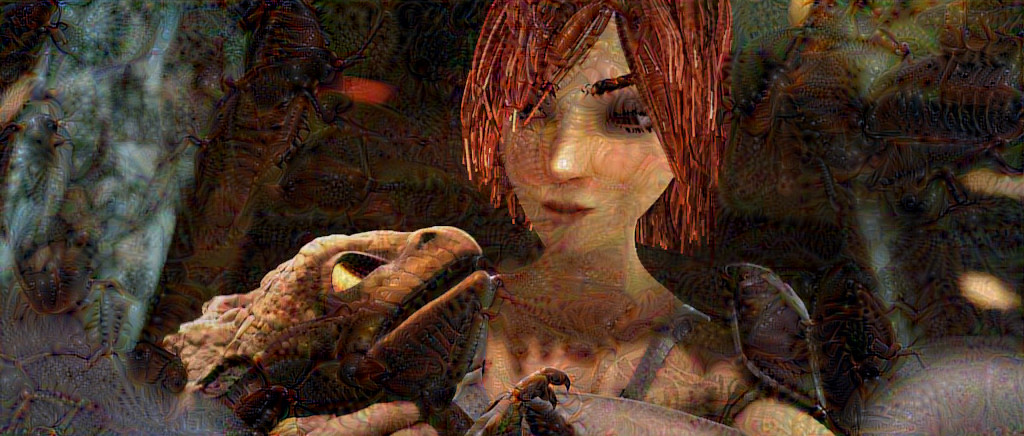}
    \includegraphics[width=.32\textwidth,height=\textheight,keepaspectratio]{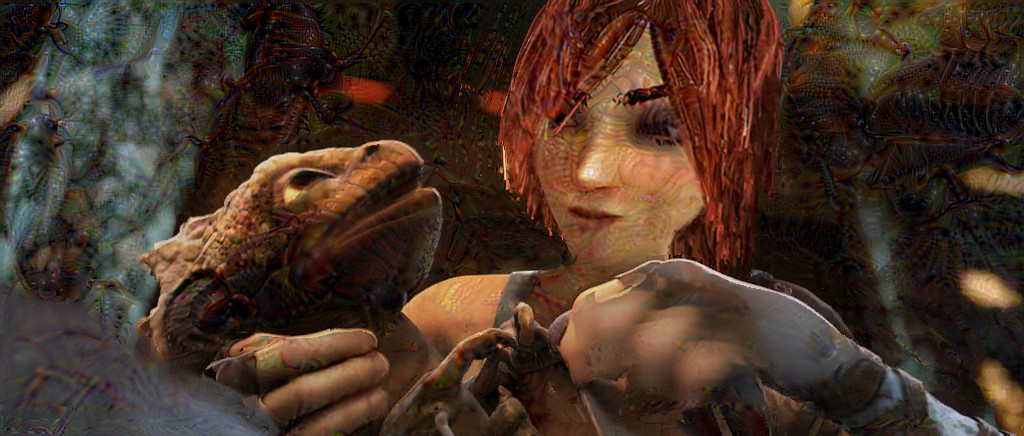}
    \includegraphics[width=.32\textwidth,height=\textheight,keepaspectratio]{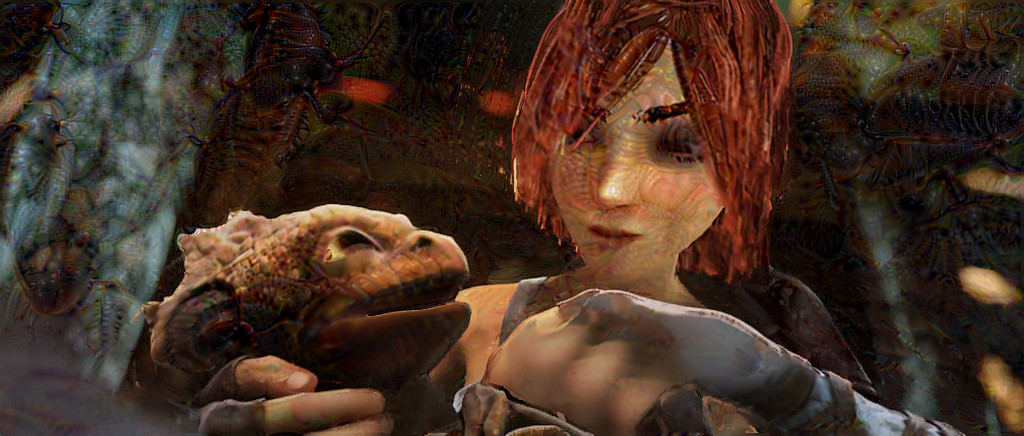}
    \includegraphics[width=.32\textwidth,height=\textheight,keepaspectratio]{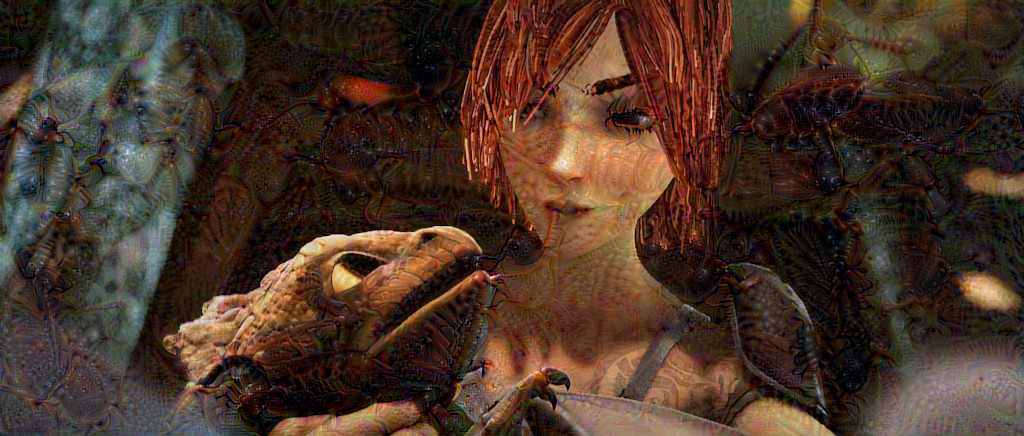}
    \includegraphics[width=.32\textwidth,height=\textheight,keepaspectratio]{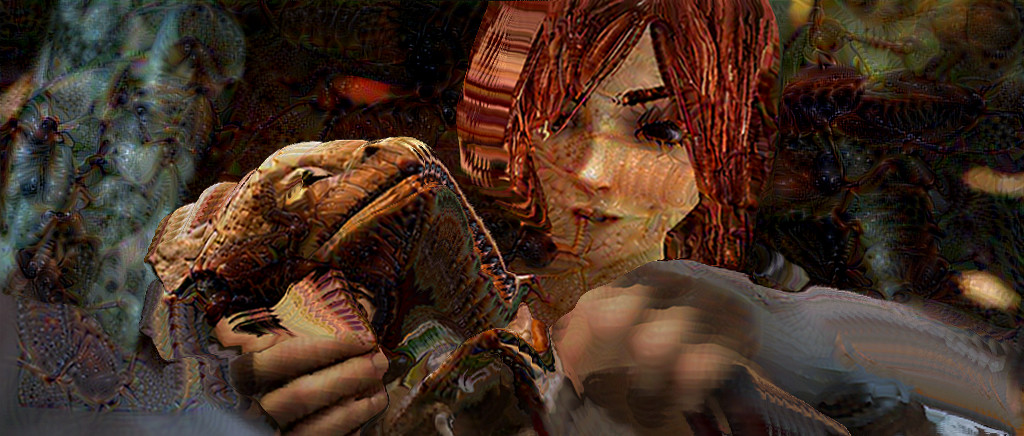}
    \includegraphics[width=.32\textwidth,height=\textheight,keepaspectratio]{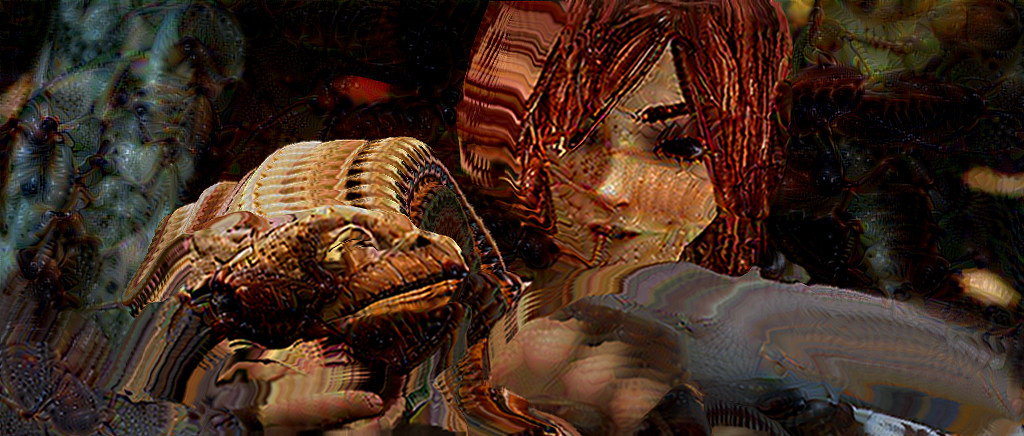}
    \includegraphics[width=.32\textwidth,height=\textheight,keepaspectratio]{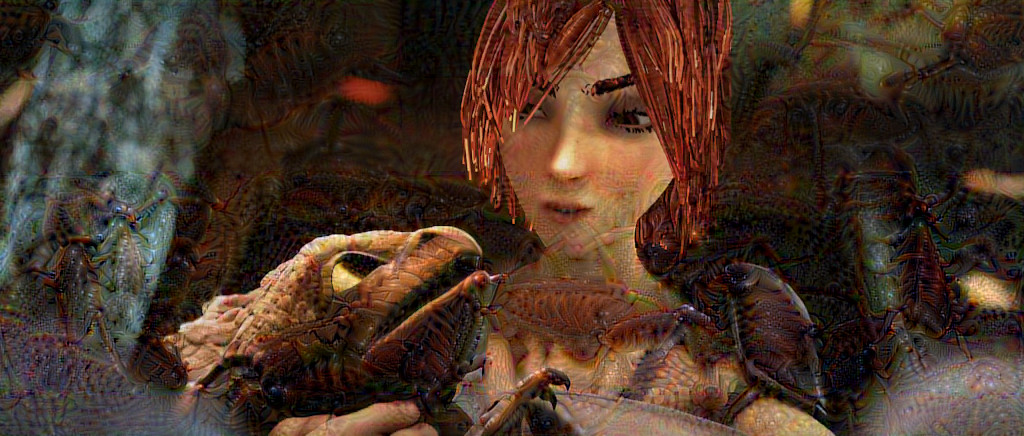}
    \includegraphics[width=.32\textwidth,height=\textheight,keepaspectratio]{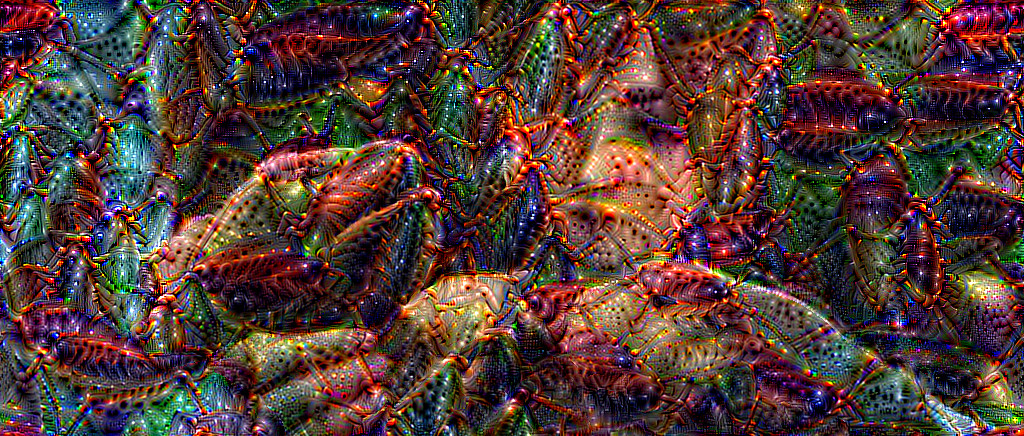}
    \includegraphics[width=.32\textwidth,height=\textheight,keepaspectratio]{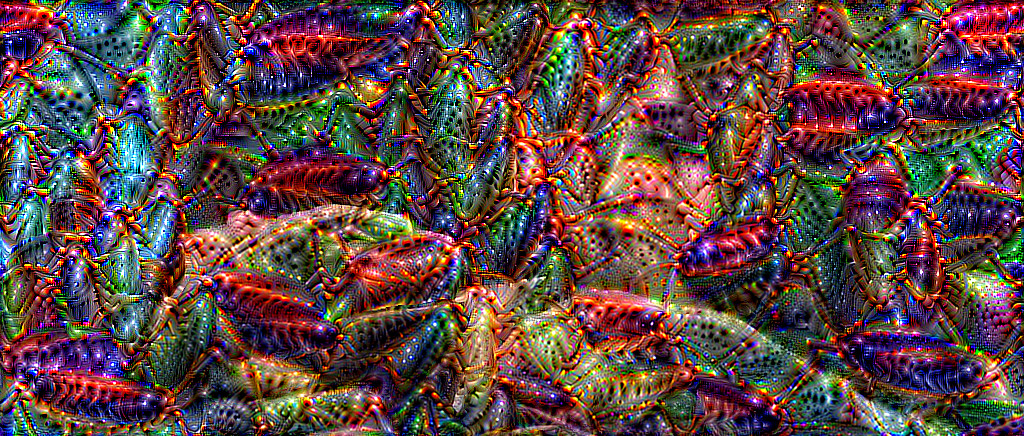}
    \includegraphics[width=.32\textwidth,height=\textheight,keepaspectratio]{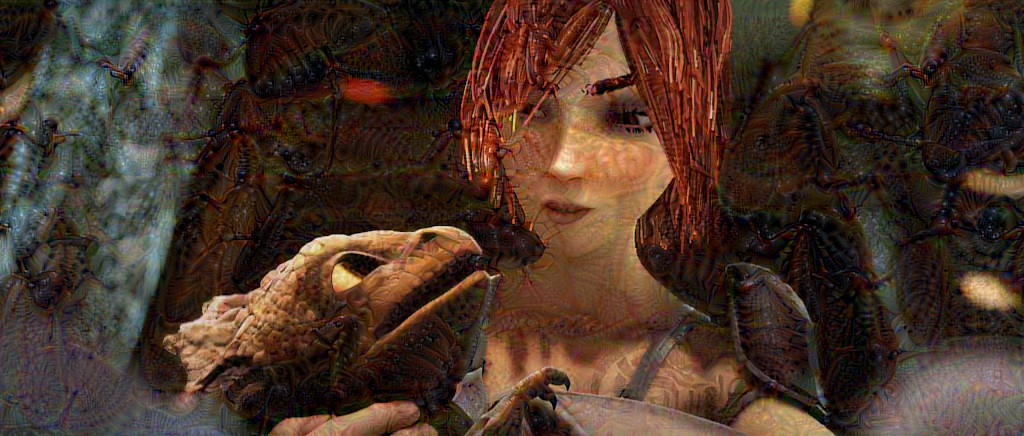}
    \includegraphics[width=.32\textwidth,height=\textheight,keepaspectratio]{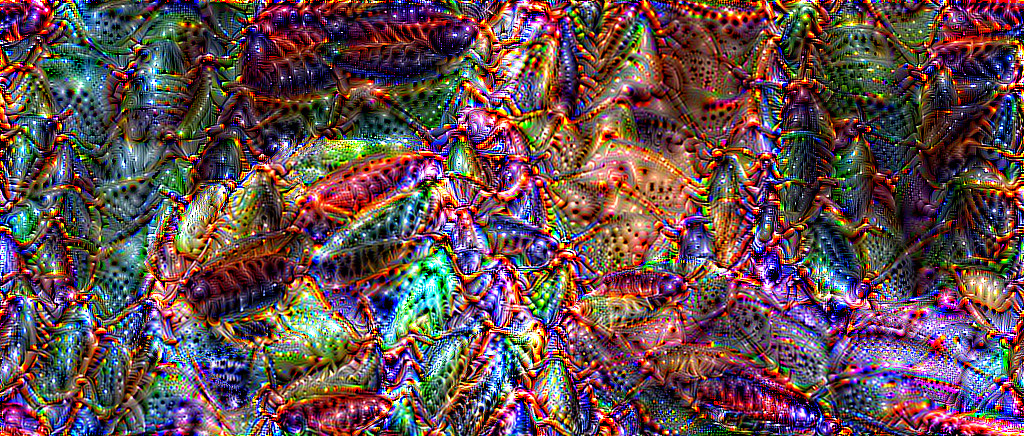}
    \includegraphics[width=.32\textwidth,height=\textheight,keepaspectratio]{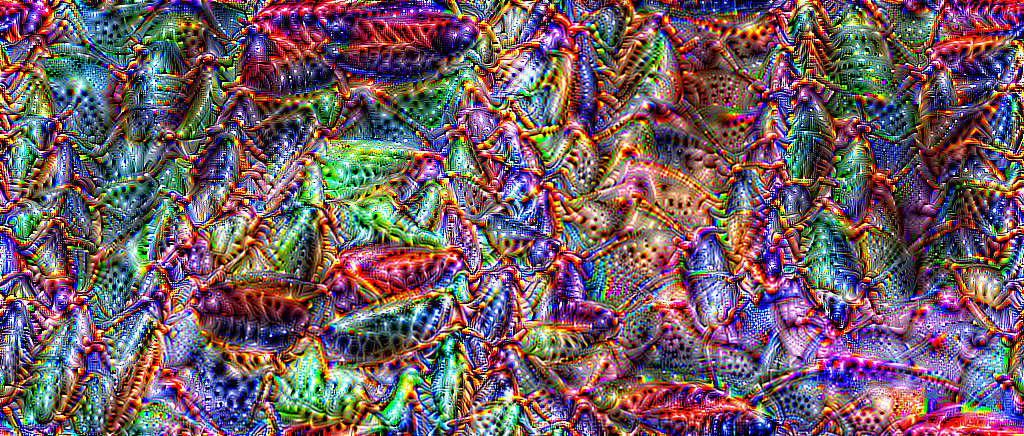}
  \caption{Column-wise, left to right: frames 1, 25 and 50 of a 50 frame video clip. Row-wise, top-down: original frames; controlled DeepDream applied frame-wise, without temporal consistency loss; controlled DeepDream with short-term temporal consistency loss;  controlled DeepDream with long-term temporal consistency loss;  controlled DeepDream with an optical flow trail;  controlled DeepDream with decay;  controlled DeepDream a flow trail and decay. \texttt{Cockroach} ImageNet class was chosen to be hallucinated, where applicable. Best viewed zoomed in and in colour. Differences are particularly clearly visible just above Scales' head, and just to the left of Sintel's head. All images of original size $1024\times436$. The videos from which these frames were taken are available \link{\href{https://drive.google.com/open?id=1SkYCJK4tmkZW7GzrMTQNt2PTkhCSz8T7}{here}}.}
  \label{fig:dragon}
\end{figure*}

During the video generation process, we find that the optical flow based loss does not work right out of the box. This is because the DeepDream loss $\mathcal{L}_{l, a}$ comprises of just one term that encourages hallucination, unlike in neural style transfer \citep{styleTransfer}, where the content loss ensures that no content in the image is lost while the style loss ensures that the style of the painting is transferred to the end result. The new objective obtained by adding in a content loss would thus be inconsistent: the content loss term aims to retain the content in the image (often measured in terms of the similarity between the activations of the input content and the output images using an L2 objective); the DeepDream loss, in direct contrast, aims to maximize the activations of a specific class or feature map-- the two loss terms thus directly contradict each other.

This simultaneous application of these conflicting optimization objectives can make some parts of the hallucinations intense, and others closer to the content. This is because grounding the content now becomes more challenging, particularly when some parts of the image are filled in from a previous iteration, but others are initialized from non-hallucinated content (for example, occluded regions that got de-occluded in the current frame). Further, the regions of the frame present in the previous frame optimize much quicker than the de-occluded regions because they are affected by both the temporal consistency and the DeepDream losses, while the temporal consistency loss remains masked out for regions corresponding to de-occlusions and edges.

We find, however, that a simple "over-hallucination" technique successfully mitigates this issue. In particular, we find it advantageous to significantly increase the number of iterations for which DeepDream is applied in all frames subsequent to the first. This over-hallucination fills in areas initialized from the original content image, making it stylistically consistent with the rest of the image initialized from the previous hallucinated frame. In spite of the increase in the number of iterations for which DeepDream is applied, we find that the over-hallucination technique does not cause subsequent frames in the video to degrade in quality by the DeepDream hallucinations becoming too intense. This is likely because the de-occluded regions are grounded by rest of the image, the rest of the image in turn being grounded by the the corresponding regions in the previous frames due to the temporal consistency loss term.

\subsection{Shot Change Detection}

A drawback of over-hallucination is that it relies on consistency with the previous frame to ground the image. If the new frame is significantly different from the previous one (as happens when there is a scene change), nothing grounds the hallucinations in the scene to their previous intensity. Consequently, the increased number of iterations cause a massive degradation in the frame, making it and subsequent frames consist of solely hallucinations with little resemblance to the original content. To avoid this, we treat poor temporal consistency between a frame and the one immediately preceding it as an indication of a shot change, applying DeepDream only for the original number of iterations on such frames (as opposed to over-hallucinating them). Empirically, we find that an $85\%$ inconsistency threshold to tag a frame as a shot change with respect to the previous frame works well.

\subsection{Implementation Details}

The first part of the proposed architecture involves obtaining optical flows in both directions, between pairs of adjacent frames when the short-term consistency loss is used, and between frames that are at the appropriate fixed distance from each other when the long-term consistency loss is used. We do this with the help of a DeepFlow2 model trained on the Sintel dataset \citep{sintel}.

To hallucinate objects in an image (or in each frame of a video) by minimizing the LucidDream loss $\mathcal{L}$, we choose to employ a pre-trained VGGNet\citep{vggnet}, pre-trained on ImageNet\citep{imagenet}. The architecture used is identical to the VGG-19 architecture, except that we use average pooling instead of max pooling, and mirror padding instead of constant-padding with zeros.

\subsection{Alternate Visual Effects} \label{sec:altvisual}

The primary objective of this work is to improve the degree of control over the class hallucinated without a complex search and to provide temporal consistency in the generated videos. However, as part of this process, we observed several other visual effects that are interesting from an artistic standpoint. In one case, we see an optical flow trail that clearly highlights the motion of objects (which we refer to as the \textbf{optical trail effect}). To achieve this, we define an additional flow-trail loss term which creates an optical flow trail that clearly highlights the motion of objects as:

\begin{equation}
    \mathcal{L}_f =  \frac{1}{D \sum_{k=1}^{D} c^{(i-j, i)}[k]} \norm{x^{(i)} - w^{(i-j, i)}}^2_\mathbf{F}
\end{equation}

This loss term is then added to the LucidDream loss; the overall loss thus becomes $\mathcal{L} + \delta\mathcal{L}_f$.

In another such visual effect, the hallucinated effect becomes progressively more intense until the entire image is filled with the object to be hallucinated (i.e., it smoothly decays the image-- referred to as the \textbf{decay effect}); a third effect combines both of these, showing an optical trail as well as progressively decaying the image (which we call \textbf{optical trail+decay}). In addition, the various visual effects when applied to a video are shown in Figure \ref{fig:dragon}, while the hyperparameters used to obtain these effects are described in Section \ref{sec:hyperparam} below.

\subsection{Hyperparameters} \label{sec:hyperparam}

We randomly select the origin as part of our update $k$ times, and update each of the so defined tiles $k$ times for each such random selection, where $k=12$ for images or frames that are the first frame or that correspond to a scene change, and $k=30$ otherwise (as part of our over-hallucination technique described in Section \ref{sec:overhall}). In the short-term case, the loss is applied only between the current frame and the previous frame; in the case of the long-term temporal consistency loss, J=\{1 2, 4, 8, 16, 32\}. In the case of the optical flow trail and optical flow trail+decay effect, we initialize each frame with the warped version of the previous hallucinated frame; in all other cases, we initialize each frame with its original content. We set $\alpha=10000$ and use the Adam optimizer \citep{kingma2014adam} throughout. In the long-term temporal loss control method, we set $\gamma=1000$ and $\beta=0$; $\gamma=0$ in all other cases. $\beta=1$ in the case of trail mode, $\beta=3$ in the case of decay and trail+decay mode, and $\beta=300$ in the case of the short-term temporal consistency mode. $\delta=0$ in all cases except the flow case, where we set $\delta=500$.

\end{document}